# Retraction and Generalized Extension of Computing with Words

Yongzhi Cao, Mingsheng Ying, and Guoqing Chen

*Abstract*— Fuzzy automata, whose input alphabet is a set of numbers or symbols, are a formal model of computing with values. Motivated by Zadeh's paradigm of computing with words rather than numbers, Ying proposed a kind of fuzzy automata, whose input alphabet consists of all fuzzy subsets of a set of symbols, as a formal model of computing with all words. In this paper, we introduce a somewhat general formal model of computing with (some special) words. The new features of the model are that the input alphabet only comprises some (not necessarily all) fuzzy subsets of a set of symbols and the fuzzy transition function can be specified arbitrarily. By employing the methodology of fuzzy control, we establish a retraction principle from computing with words to computing with values for handling crisp inputs and a generalized extension principle from computing with words to computing with all words for handling fuzzy inputs. These principles show that computing with values and computing with all words can be respectively implemented by computing with words. Some algebraic properties of retractions and generalized extensions are addressed as well.

## I. INTRODUCTION

ONE of the most remarkable capabilities of the human is the capability of performing a wide variety of physical and mental tasks by using perceptions in purposeful ways and approximating perceptions via propositions in natural language. Thus, developing perception-based machines may be a feasible approach to constructing intelligent systems. Motivated by this, Zadeh proposed and advocated the idea of computing with words in a series of papers [33]–[37]. Computing with words tries to present a conceptual framework for computing and reasoning with words rather than numbers, where words play the role of labels of perceptions. Since its introduction, the concept of computing with words has gained considerable attention, including some successful applications in information processing, decision, and control [9], [10], [14], [27], [28], [38].

Computing, in its traditional sense, is centered on manipulation of numbers and symbols, and is usually represented by a dynamic model in which an input device is equipped. In contrast, computing with words is a methodology in which the objects of computation are words and propositions drawn from a natural language. It is worth noting that most of the literature on computing with words is devoted to developing new computationally feasible algorithms for uncertain reasoning. In these works, the word "computing" in the phrase "computing with words" means computational implementations of uncertain reasoning; it is irrelevant to the formal theory of computing.

In his paper [30], Ying incorporated the initial idea of computing with words together with classical models of computation and then proposed a formal model of computing with words in terms of fuzzy automata. It is well known that automata are the prime example of general computational systems. In an automaton, the input alphabet consists of a finite number of discrete input symbols. These input symbols may be reasonably thought of as the input values that we are going to compute. Fuzzy automata initiated by Santos [23] resulted from combining automata theory with fuzzy logic, in which state transitions are imprecise and uncertain. It is this property that makes it possible to model uncertainty which is inherent in many applications (see, for example, [5], [11], and [15]). Nevertheless, the input alphabet of a fuzzy automaton appearing in the literature on computation is the same as that of an automaton, although a certain impreciseness or uncertainty is involved in the process of computation. Consequently, these fuzzy automata can still be thought of as models of computing with values. The key idea underlying Ying's formal model of computing with words is the use of words in place of values as input symbols of a fuzzy automaton, where words are formally represented as fuzzy subsets of the input alphabet, i.e., possibility distributions over the input alphabet. Such an idea has been developed by Qiu and Wang in [21] and [24].

Following [30], we identify a value with a symbol from the input alphabet and also a word with a fuzzy subset of the input alphabet, and use them exchangeably. For clarity, it is convenient to name three kinds of fuzzy automata explicitly. Informally, we say a fuzzy automaton is a fuzzy automaton for computing with values (FACV) if its input alphabet is a finite set of symbols (values). A fuzzy automaton is said to be a fuzzy automaton for computing with words (FACW) if its input alphabet consists of some (not necessarily all) fuzzy subsets of a finite set of symbols; in particular, the fuzzy automaton is called a fuzzy automaton for computing with all words (FACAW) if its input alphabet consists of all fuzzy subsets of the set of symbols. The point of departure in [30] is an FACV. By exploiting extension, the FACV gives rise to an FACAW that models formally computing with words. This process of obtaining a formal model for computing with

This work was supported by the National Foundation of Natural Sciences of China under Grants 60505011, 60496321, 70231010, and 70321001, and by the Chinese National Key Foundation Research & Development Plan (2004CB318108).

Y. Z. Cao and G. Q. Chen are with the School of Economics and Management, Tsinghua University, Beijing 100084, China (e-mail: caoyz@mail.tsinghua.edu.cn, chengq@em.tsinghua.edu.cn).

M. S. Ying is with the State Key Laboratory of Intelligent Technology and Systems, Department of Computer Science and Technology, Tsinghua University, Beijing 100084, China (e-mail: yingmsh@mail.tsinghua.edu.cn).

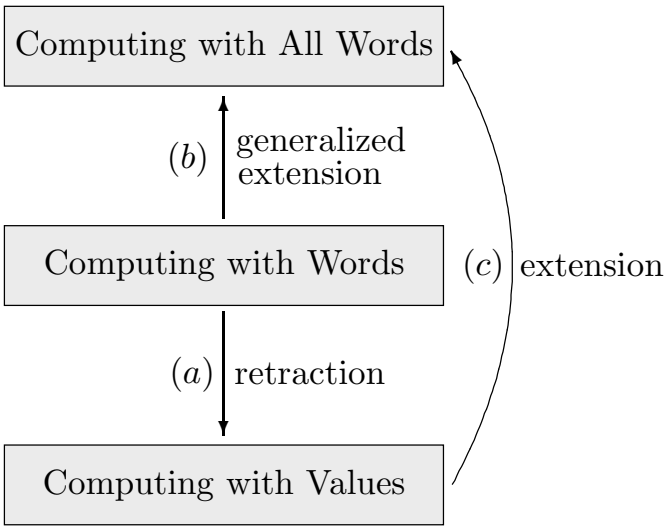

Fig. 1. Interrelation among retractions, extensions, and generalized extensions.

words just corresponds to the extension $(c)$ in Fig. 1. Roughly speaking, Ying's formal model of computing with words is an FACAW arising from a certain FACV. As a consequence, the FACAW inevitably depends on the underlying FACV. This observation motivates us to propose a somewhat general formal model of computing with words.

In this paper, we adopt FACWs to model formally computing with words. The new features of the model are that the input alphabet consists of some (not necessarily all) fuzzy subsets and the fuzzy transition function can be specified arbitrarily. After introducing this model, we embark upon two directional extensions: retractions and generalized extensions. The start point of the extensions is an FACW modeling computing with (some special) words. For the retraction, we establish a retraction principle from computing with words to computing with values which means that computing with values can be implemented by computing with words with the price of a large number of extra computations; for the generalized extension, we establish a generalized extension principle from computing with words to computing with all words which means that computing with all words can be implemented by computing with words with the price of a large number of extra computations. In fact, the generalized extension provides an interpolation approach which helps reduce the complexity of devising a model for computing with words.

The motivation behind retractions and generalized extensions comes from fuzzy control theory, where the underlying fuzzy logic system is used to deal with crisp inputs as well as fuzzy inputs by different fuzzifications. Both the retraction principle and the generalized extension principle here are derived by employing the methodology of fuzzy control (see, e.g., [7], [17], [25]). The purpose of developing the principles in this way is twofold. Firstly, it aims to provide a background semantics in doing so. Secondly, it aims to give a paradigm for establishing other extension principles by using other fuzzifications and inference engines which have been intensively studied in the field of fuzzy control (see Chapters 7 and 8 of [25] for some examples).

As a result, the retraction that corresponds to the process $(a)$ in Fig. 1 yields an FACV for handling crisp inputs and the generalized extension that corresponds to the process $(b)$ in Fig. 1 yields an FACAW for handling fuzzy inputs. By identifying each symbol from the input alphabet with a singleton, each FACV can be viewed as an FACW. In this sense, we see later that the process $(c)$ in Fig. 1 is a special case of the process $(b)$, which justifies the name of generalized extension on the one hand. On the other hand, the name is also justified by the fact that the generalized extension is not an extension in a strictly mathematical sense which will be made precise. Further, we will verify that the process $(b)$ in Fig. 1 produces the same outcome as that of the composition of processes $(a)$ and $(c)$. Finally, in order to render retractions and generalized extensions more useful for real-world applications (for example, large-scale systems), we turn to investigating the preservation property of retractions and generalized extensions under the use of product operators and homomorphisms.

The remainder of this paper is structured as follows. In Section II, we briefly review some basics of fuzzy automata and introduce the formal model of computing with words. Sections III and IV are devoted to retractions and generalized extensions, respectively. Some relationships among retractions, extensions, and generalized extensions are explored in Section V. We discuss certain algebraic properties of retractions and generalized extensions in Section VI and conclude the paper in Section VII. The proofs of our theorems are given in Appendix I.

## II. FORMAL MODEL OF COMPUTING WITH WORDS

To introduce a formal model of computing with words, let us first review some notions on fuzzy set theory and fuzzy automata. For a detailed introduction to the notions, the reader may refer to [12], [18], and [15].

Let $X$ be a universal set. A *fuzzy set* $A$ [31], or rather a *fuzzy subset* $A$ of $X$, is defined by a function assigning to each element $x$ of $X$ a value $A(x)$ in the closed unit interval $[0, 1]$. Such a function is called a *membership function*, which is a generalization of the characteristic function associated to a crisp set; the value $A(x)$ represents the membership grade of $x$ in $A$, which characterizes the degree of membership of $x$ in $A$.

We denote by $\mathcal{F}(X)$ the set of all fuzzy subsets of $X$. For any $A, B \in \mathcal{F}(X)$, we say that $A$ is contained in $B$ (or $B$ contains $A$), denoted by $A \subseteq B$, if $A(x) \leq B(x)$ for all $x \in X$. We say that $A = B$ if and only if $A \subseteq B$ and $B \subseteq A$. A fuzzy set is said to be *empty* if its membership function is identically zero on $X$. We use $\emptyset$ to denote the empty fuzzy set.

The *support* of a fuzzy set $A$ is a crisp set defined as $\mathrm{supp}(A) = \{x \in X : A(x) > 0\}$. Whenever $\mathrm{supp}(A)$ is a finite set, say $\mathrm{supp}(A) = \{x_1, x_2, \ldots, x_n\}$, we may write $A$ in Zadeh's notation as
$$A = \frac{A(x_1)}{x_1} + \frac{A(x_2)}{x_2} + \cdots + \frac{A(x_n)}{x_n}.$$

For any family $\lambda_i$, $i \in I$, of elements of $[0,1]$, we write $\vee_{i \in I} \lambda_i$ or $\vee\{\lambda_i : i \in I\}$ for the supremum of $\{\lambda_i : i \in I\}$, and $\wedge_{i \in I} \lambda_i$ or $\wedge\{\lambda_i : i \in I\}$ for the infimum. In particular, if $I$ is finite, then $\vee_{i \in I} \lambda_i$ and $\wedge_{i \in I} \lambda_i$ are the greatest element and the least element of $\{\lambda_i : i \in I\}$, respectively. For any $A \in \mathcal{F}(X)$, the *height* of $A$ is defined as

$$\text{height}(A) = \vee_{x \in X} A(x).$$

Given $A, B \in \mathcal{F}(X)$, the *union* of $A$ and $B$, denoted $A \cup B$, is defined by the membership function

$$(A \cup B)(x) = A(x) \vee B(x)$$

for all $x \in X$; the *intersection* of $A$ and $B$, denoted $A \cap B$, is given by the membership function

$$(A \cap B)(x) = A(x) \wedge B(x)$$

for all $x \in X$. Let $\lambda \in [0,1]$ and $A \in \mathcal{F}(X)$. The *scale product* $\lambda \cdot A$ of $\lambda$ and $A$ is defined by

$$(\lambda \cdot A)(x) = \lambda \wedge A(x)$$

for every $x \in X$; this is again a fuzzy subset of $X$.

For later need, let us recall Zadeh's extension principle. If $X$ and $Y$ are two crisp sets and $f$ is a mapping from $X$ to $Y$, then $f$ can be extended to a mapping from $\mathcal{F}(X)$ to $\mathcal{F}(Y)$ in the following way: For any $A \in \mathcal{F}(X)$, $f(A) \in \mathcal{F}(Y)$ is given by

$$f(A)(y) = \vee\{A(x) : x \in X \text{ and } f(x) = y\}$$

for all $y \in Y$.

We are ready to review the concepts of fuzzy automaton and fuzzy language. The fuzzy automata here have been known as max-min automata in some mathematical literature [11], [23].

*Definition 1:* A *fuzzy automaton* is a five-tuple $M = (Q, \Sigma, \delta, q_0, F)$, where:

1) $Q$ is a finite set of states.
2) $\Sigma$ is a finite input alphabet.
3) $q_0$, a member of $Q$, is the initial state.
4) $F$ is a fuzzy subset of $Q$, called the fuzzy set of final states and for each $q \in Q$, $F(q)$ indicates intuitively the degree to which $q$ is a final state.
5) $\delta$, the fuzzy transition function, is a function from $Q \times \Sigma$ to $\mathcal{F}(Q)$ that takes a state in $Q$ and an input symbol in $\Sigma$ as arguments and returns a fuzzy subset of $Q$.

For any $p, q \in Q$ and $a \in \Sigma$, we can interpret $\delta(p,a)(q)$ as the possibility degree to which the automaton in state $p$ and with input $a$ may enter state $q$.

Denote by $\Sigma^*$ the set of all finite strings constructed by concatenation of elements of $\Sigma$, including the empty string $\epsilon$. In the literature of classical automata theory, a string is often called a "word". Like [30], to avoid confusion in this paper, we do not use the term "word" in this way and only use it to refer to what we mean by "word" in the phrase "computing with words."

To describe what happens when we start in any state and follow any sequence of inputs, we extend the fuzzy transition function to strings.

*Definition 2:* Let $M = (Q, \Sigma, \delta, q_0, F)$ be a fuzzy automaton.

1) The *extended fuzzy transition function* from $Q \times \Sigma^*$ to $\mathcal{F}(Q)$, denoted by the same notation $\delta$, is defined inductively as follows:

$$\begin{aligned} \delta(p, \epsilon) &= \frac{1}{p} \\ \delta(p, wa) &= \cup_{q \in Q} [\delta(p, w)(q) \cdot \delta(q, a)] \end{aligned}$$

for all $w \in \Sigma^*$ and $a \in \Sigma$, where $1/p$ is a singleton in $Q$, i.e., the fuzzy subset of $Q$ with membership 1 at $p$ and with zero membership for all the other elements of $Q$. In addition, $\delta(p, w)(q) \cdot \delta(q, a)$ stands for the scale product of the membership $\delta(p, w)(q)$ and the fuzzy set $\delta(q, a)$.

2) The *language $L(M)$ accepted by $M$* is a fuzzy subset of $\Sigma^*$ with the membership function defined by

$$L(M)(w) = \text{height}(\delta(q_0, w) \cap F)$$

for all $w \in \Sigma^*$. The membership $L(M)(w)$ is the degree to which $w$ is accepted by $M$.

The above definitions provide a model of computing with values on fuzzy automata. We shall refer to the fuzzy automaton in Definition 1 as a *fuzzy automaton for computing with values* (or FACV for short). Within the framework of fuzzy automata, Ying [30] proposed the following formal model of computing with words by extending further the fuzzy transition function of an FACV.

*Definition 3:* Let $M = (Q, \Sigma, \delta, q_0, F)$ be an FACV.

1) To deal with words as inputs, $\delta$ is extended to a function from $Q \times \mathcal{F}(\Sigma)$ to $\mathcal{F}(Q)$, which is denoted by $\hat{\delta}$, with Zadeh's extension principle:

$$\hat{\delta}(p, A) = \cup_{a \in \Sigma} [A(a) \cdot \delta(p, a)]$$

for any $p \in Q$ and $A \in \mathcal{F}(\Sigma)$. This gives rise to a fuzzy automaton $\hat{M} = (Q, \mathcal{F}(\Sigma), \hat{\delta}, q_0, F)$.

2) To deal with strings of words as inputs, $\hat{\delta}$ in 1) is further extended to a function from $Q \times \mathcal{F}(\Sigma)^*$ to $\mathcal{F}(Q)$, denoted again by $\hat{\delta}$, as follows:

$$\begin{aligned} \hat{\delta}(p, \epsilon) &= \frac{1}{p} \\ \hat{\delta}(p, WA) &= \cup_{q \in Q} [\hat{\delta}(p, W)(q) \cdot \hat{\delta}(q, A)] \end{aligned}$$

for all $W \in \mathcal{F}(\Sigma)^*$ and $A \in \mathcal{F}(\Sigma)$.

3) The *word language $L_w(\hat{M})$ accepted by $\hat{M}$* is a fuzzy subset of $\mathcal{F}(\Sigma)^*$ with the membership function defined by

$$L_w(\hat{M})(W) = \text{height}(\hat{\delta}(q_0, W) \cap F)$$

for all $W \in \mathcal{F}(\Sigma)^*$. The membership $L_w(\hat{M})(W)$ is the degree to which the string $W$ of words is accepted by $\hat{M}$.

As we see from Definition 3, the formal model of computing with words proposed by Ying is dynamic in the sense that the output generally depends on past words of the input, and it is essentially a fuzzy automaton that is the same as the fuzzy automaton in Definition 2. Importantly, however, the strings of inputs are different: in Definition 2 they are strings of values, whereas in Definition 3 they are strings of words. It is worth noting that the input alphabet of $\hat{M}$ consists of all fuzzy subsets of $\Sigma$ and the fuzzy transition function $\hat{\delta}$ in 1) of Definition 3 is also dependent on the underlying fuzzy automaton $M$. For this reason, we introduce a somewhat general model of computing with words.

*Definition 4:* A *fuzzy automaton for computing with words* (or FACW for short) is a fuzzy automaton $\tilde{M} = (Q, \tilde{\Sigma}, \tilde{\delta}, q_0, F)$, where the components $Q, q_0, F$ have their same interpretation as in Definition 1 and the following hold:

2)' $\tilde{\Sigma}$ is a subset of $\mathcal{F}(\Sigma)$, where $\Sigma$ is a finite set of symbols, called the underlying input alphabet.
5)' $\tilde{\delta}$ is a fuzzy transition function from $Q \times \tilde{\Sigma}$ to $\mathcal{F}(Q)$.

The new features of the model in Definition 4 are that the input alphabet consists of some (not necessarily all) fuzzy subsets of a finite set of symbols (i.e., the underlying input alphabet) and the fuzzy transition function can be specified arbitrarily. In particular, when $\tilde{\Sigma} = \mathcal{F}(\Sigma)$, we say that the FACW is a *fuzzy automaton for computing with all words* (or FACAW for short). By and large, the choice of $\tilde{\Sigma}$ and the specification of the fuzzy transition function $\tilde{\delta}$ are subjective; they are provided by the expert in an ad hoc (heuristic) manner from experience or intuition. Nevertheless, in order to provide more information for computing with values and computing with all words, it is better to choose suitably more words for $\tilde{\Sigma}$ and require $\tilde{\Sigma}$ to be *complete*. (The completeness means that for each $a \in \Sigma$, there exists a word $A \in \tilde{\Sigma}$ such that $A(a) > 0$.) There are also some methods (see, for example, pages 256-260 of [6] and pages 19-26 of [18]) for estimating the fuzzy transition function $\tilde{\delta}$ that determines by some membership functions. Definition 2 is applicable to FACWs, and we thus get a direct way of computing the string of words.

*Example 1:* Let us assume that we are considering a gas cooker. The temperature of the gas cooker is experientially classified as three states: $q_0, q_1$, and $q_2$, where $q_0, q_1, q_2$ represent "low", "medium", and "high", respectively. We consider the flux of gas as inputs, which is described by linguistic expressions (namely, words): $S = small$, $M = medium$, and $L = large$. More explicitly, these words interpreted as fuzzy sets are defined as follows:

$$S = small = \frac{1}{1} + \frac{0.5}{2} + \frac{0.1}{3},$$
$$M = medium = \frac{0.2}{2} + \frac{1}{3} + \frac{0.2}{4},$$
$$L = large = \frac{0.1}{3} + \frac{0.5}{4} + \frac{1}{5},$$

where the underlying input alphabet $\Sigma$ consists of discretized flux, i.e., $\Sigma = \{1,2,3,4,5\}$. We take $q_0$ as the initial state and $F = 0.1/q_0 + 1/q_1 + 0.1/q_2$ as the fuzzy set of final states. The fuzzy transition function $\tilde{\delta}$ is depicted in Fig. 2, where an arc from $q_i$ to $q_j$ with label $W|x$ means that $\tilde{\delta}(q_i, W)(q_j) = x$. A box with $q_i|y$ in Fig. 2 means that $F(q_i) = y$. We thus get an FACW $\tilde{M} = (Q = \{q_0, q_1, q_2\}, \tilde{\Sigma} = \{S, M, L\}, \tilde{\delta}, q_0, F)$. According to the fuzzy transition function, we can compute the word language accepted by $\tilde{M}$. For example, the degree to which the string "$SLM$" is accepted is $0.7$.

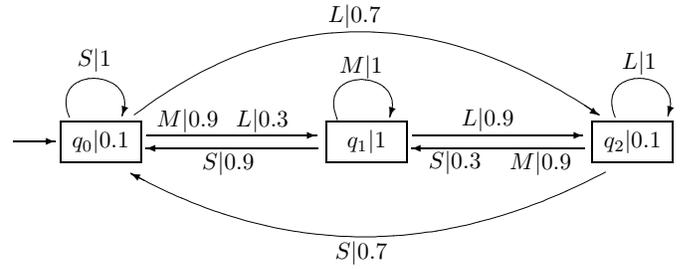

Fig. 2. An FACW modeling the relationship between the temperature and the gas flux of a gas cooker.

## III. RETRACTIONS: COMPUTING WITH VALUES

Recall that the formal model derived by Ying is in fact an extension from computing with values to computing with words. In this section, we in turn address how to tackle computing with values when we only have a fuzzy automaton $\tilde{M} = (Q, \tilde{\Sigma}, \tilde{\delta}, q_0, F)$ for computing with words. To this end, we shall establish a fuzzy automaton $\tilde{M}^{\downarrow} = (Q, \Sigma, \delta^{\downarrow}, q_0, F)$, where the components $Q, q_0, F$ are the same as those of $\tilde{M}$, $\Sigma$ is the underlying input alphabet of $\tilde{M}$, and $\delta^{\downarrow}$ which depends on the fuzzy transition function of $\tilde{M}$ need to be defined.

As mentioned in the introduction, we will define $\delta^{\downarrow}$ by using the methodology of fuzzy control (see, e.g., [17] and [25]). To do so, let us first recall the general scheme of a fuzzy logic system based on fuzzy IF-THEN rules. A fuzzy IF-THEN rule expresses a fuzzy implication relation between the fuzzy sets of the premise and the fuzzy sets of the conclusion. Each fuzzy IF-THEN rule is of the following form

IF premise THEN consequent.

For example, a fuzzy IF-THEN rule for controlling a product line could be

IF *buffer level* is *Full* AND *surplus* is *Positive*, THEN *production rate* is *Low*,

where *Full*, *Positive*, and *Low* are fuzzy sets defined on corresponding universal sets. This is a *condition $\rightarrow$ action* rule, which means that the production rate should be decreased if the buffer is full and, moreover, the production is superfluous.

A *fuzzy logic system* (FLS) is comprised of four components:

- Rule-base: It consists of some fuzzy IF-THEN rules.
- Fuzzification: It encodes the crisp inputs into fuzzy sets described by linguistic expressions.
- Inference engine: It uses the fuzzy rules in the rule-base to produce fuzzy conclusions (e.g., the implied fuzzy sets).
- Defuzzification: It decodes the inferred fuzzy conclusions into crisp outputs.

For our purpose of defining $\delta^{\downarrow}$, there is no need for developing a complete FLS; in fact, the defuzzification component will not be considered.

Let $\tilde{M} = (Q, \tilde{\Sigma}, \tilde{\delta}, q_0, F)$ be an FACW. Recall that if the current state is $p$ and the input is $A$, then by definition the next state distribution is $\tilde{\delta}(p, A)$. According to this, we associate to each pair of $p \in Q$ and $A \in \tilde{\Sigma}$ a fuzzy IF-THEN rule $\mathrm{R}_p^A$:

IF *current state* is $\frac{1}{p}$ AND *input* is $A$, THEN *next state distribution* is $\tilde{\delta}(p, A)$.

The rule means that the possibility that the next state is $q$ is exactly $\tilde{\delta}(p, A)(q)$, if the current state is $p$ and the input is $A$. The rule-base associated to $\tilde{M}$, denoted by $\mathfrak{R}$, consists of all such fuzzy IF-THEN rules $\mathrm{R}_p^A$. If $\tilde{\delta}(p, A) = \emptyset$ for some $p \in Q$ and $A \in \tilde{\Sigma}$, then we exclude the fuzzy IF-THEN rule $\mathrm{R}_p^A$ from the rule-base $\mathfrak{R}$ since, as we will see later, this rule does not contribute to the computation of $\delta^\downarrow$. So there are at most $|Q| \cdot |\tilde{\Sigma}|$ rules in $\mathfrak{R}$, where the notation $|X|$ denotes the cardinality of $X$.

Having built the rule-base, we now turn to the fuzzification. This process is to specify how the FLS will convert its crisp inputs into fuzzy sets that are used to quantify the information in the rule-base. Formally, for a universal set $X$ and any $x \in X$, fuzzification transforms $x$ to a fuzzy subset of $X$, denoted by $\tilde{x}$; in other words, fuzzification is a mapping from $X$ to $\mathcal{F}(X)$. Quite often *singleton fuzzification* is used in the fuzzy control community, which produces a singleton fuzzy set $\tilde{x} = 1/x$. Note that the singleton fuzzy set is nothing other than a different representation for an element of the universal set, so we sometimes identify $x \in X$ with $\tilde{x} = 1/x$ when this is convenient.

The inference engine has two basic tasks: The first is to determine the extent to which each fuzzy IF-THEN rule in $\mathfrak{R}$ is relevant to the current situation characterized by the current state and the input (we call this task *matching*); the second is to draw conclusions by using the information in the rule-base that relates to the current situation (we call this task an *inference step*).

Suppose that the current state of $\tilde{M}^\downarrow$ is $q \in Q$ and the input of $\tilde{M}^\downarrow$ is $a \in \Sigma$. Using singleton fuzzification, we get two fuzzy sets $\tilde{q}$ and $\tilde{a}$ that respectively describe the current state and the input of the FLS. Following the standard approach used in fuzzy control, for any fuzzy IF-THEN rule $\mathrm{R}_p^A$, let us define two fuzzy sets $S_p^A \in \mathcal{F}(Q)$ and $I_p^A \in \mathcal{F}(\Sigma)$ as follows:

$$S_p^A(q) = (\frac{1}{p} \cap \tilde{q})(q) \text{ for any } q \in Q;$$

$$I_p^A(a) = (A \cap \tilde{a})(a) \text{ for any } a \in \Sigma.$$

The membership value $S_p^A(q)$ characterizes the matching degree between the current state $\tilde{q}$ and the rule premise $1/p$ of $\mathrm{R}_p^A$; similarly, $I_p^A(a)$ characterizes the matching degree between the input $\tilde{a}$ and the rule premise $A$ of $\mathrm{R}_p^A$. By a straightforward calculation, we find that

$$S_p^A = \frac{1}{p} \text{ and } I_p^A = A.$$

Further, for the fuzzy IF-THEN rule $\mathrm{R}_p^A$ we define a fuzzy set $M_p^A \in \mathcal{F}(Q \times \Sigma)$ with the membership function given by

$$M_p^A(q, a) = S_p^A(q) \wedge I_p^A(a).$$

Intuitively, $M_p^A(q, a)$ represents the certainty that the premise of $\mathrm{R}_p^A$ holds for the current state $q$ and input $a$ when we use singleton fuzzification. By the previous argument, we see that

$$M_p^A(q, a) = \frac{1}{p}(q) \wedge A(a),$$

that is,

$$M_p^A(q, a) = \begin{cases} A(a), & \text{if } q = p \\ 0, & \text{otherwise.} \end{cases}$$

Since $p$ and $A$ are arbitrary, this concludes the process of matching the current state $q$ and crisp input $a$ with the premises of the rules.

Let us now turn to the inference step. For each rule $\mathrm{R}_p^A$, we define an *implied fuzzy set*, denoted $N_p^A$, with the membership function

$$N_p^A(q') = M_p^A(q, a) \wedge \tilde{\delta}(p, A)(q')$$

for any $q' \in Q$. The implied fuzzy set $N_p^A$ specifies the certainty level that the next state should be $q'$, taking into consideration merely the rule $\mathrm{R}_p^A$. The *overall implied fuzzy set*, denoted $\bar{N}$, is given by the following membership function

$$\bar{N}(q') = \vee_{(p,A) \in Q \times \tilde{\Sigma}} N_p^A(q')$$

for any $q' \in Q$. The membership value $\bar{N}(q')$ is a result of considering all the rules in the rule-base at the same time, and gives the maximal possibility degree of $q'$ as the next state when the current state is $q$ and the input is $a$.

It follows from the above inference step that it is rational to define $\delta^\downarrow(q, a) = \bar{N}$. We can present an explicit expression for the membership function of $\delta^\downarrow(q, a)$ by the following computation.

$$\begin{aligned} \delta^\downarrow(q, a)(q') &= \bar{N}(q') \\ &= \vee_{(p,A) \in Q \times \tilde{\Sigma}} N_p^A(q') \\ &= \vee_{(p,A) \in Q \times \tilde{\Sigma}} [M_p^A(q, a) \wedge \tilde{\delta}(p, A)(q')] \\ &= \vee_{(q,A) \in \{q\} \times \tilde{\Sigma}} [M_q^A(q, a) \wedge \tilde{\delta}(q, A)(q')] \\ &= \vee_{A \in \tilde{\Sigma}} [A(a) \wedge \tilde{\delta}(q, A)(q')], \end{aligned}$$

i.e.,

$$\delta^\downarrow(q, a)(q') = \vee_{A \in \tilde{\Sigma}} [A(a) \wedge \tilde{\delta}(q, A)(q')]$$

for any $q' \in Q$. From the computation, we also see that $\tilde{\delta}(p, A)$ has no contribution to $\delta^\downarrow(q, a)$ if $\tilde{\delta}(p, A) = \emptyset$.

Finally, we have the following definition.

*Definition 5:* Let $\tilde{M} = (Q, \tilde{\Sigma}, \tilde{\delta}, q_0, F)$ be an FACW. The *retraction* of $\tilde{M}$ is an FACV $\tilde{M}^\downarrow = (Q, \Sigma, \delta^\downarrow, q_0, F)$, where the components $Q, q_0, F$ are the same as those of $\tilde{M}$, $\Sigma$ is the underlying input alphabet of $\tilde{M}$, and $\delta^\downarrow$ is a mapping from $Q \times \Sigma$ to $\mathcal{F}(Q)$ that maps $(q, a) \in Q \times \Sigma$ to a fuzzy subset $\delta^\downarrow(q, a)$ of $Q$ with the membership function

$$\delta^\downarrow(q, a)(q') = \vee_{A \in \tilde{\Sigma}} [A(a) \wedge \tilde{\delta}(q, A)(q')]$$

for any $q' \in Q$.

The retraction of $\tilde{M}$ deals with crisp inputs, and thus it may serve as a device for computing with values. We will refer to "↓" as the operation of obtaining the retraction. As an example, we derive the retraction of the FACW given in Example 1.

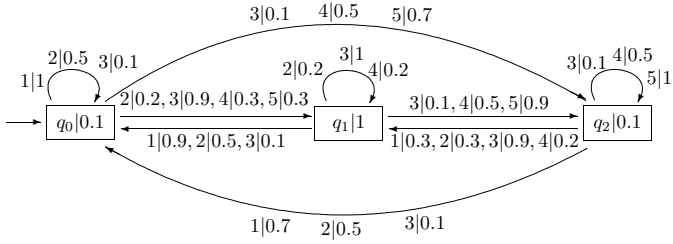

Fig. 3. Retraction of the FACW given in Example 1.

*Example 2:* Consider the FACW $\tilde{M}$ given in Example 1. By Definition 5, a straightforward calculation yields that the retraction of $\tilde{M}$ is $\tilde{M}^{\downarrow} = (Q, \Sigma = \{1,2,3,4,5\}, \delta^{\downarrow}, q_0, F)$, where the components $Q, q_0, F$ are the same as those in Example 1 and $\delta^{\downarrow}$ is depicted in Fig. 3. We compute the membership grades of $\delta^{\downarrow}(q_0, 3)$ as an example:

$$\begin{aligned}
\delta^{\downarrow}(q_0,3)(q_0) &= [S(3) \wedge \tilde{\delta}(q_0,S)(q_0)] \\
&\quad \vee [M(3) \wedge \tilde{\delta}(q_0,M)(q_0)] \\
&\quad \vee [L(3) \wedge \tilde{\delta}(q_0,L)(q_0)] \\
&= (0.1 \wedge 1) \vee (1 \wedge 0) \vee (0.1 \wedge 0) \\
&= 0.1, \\
\delta^{\downarrow}(q_0,3)(q_1) &= [S(3) \wedge \tilde{\delta}(q_0,S)(q_1)] \\
&\quad \vee [M(3) \wedge \tilde{\delta}(q_0,M)(q_1)] \\
&\quad \vee [L(3) \wedge \tilde{\delta}(q_0,L)(q_1)] \\
&= (0.1 \wedge 0) \vee (1 \wedge 0.9) \vee (0.1 \wedge 0.3) \\
&= 0.9, \\
\delta^{\downarrow}(q_0,3)(q_2) &= [S(3) \wedge \tilde{\delta}(q_0,S)(q_2)] \\
&\quad \vee [M(3) \wedge \tilde{\delta}(q_0,M)(q_2)] \\
&\quad \vee [L(3) \wedge \tilde{\delta}(q_0,L)(q_2)] \\
&= (0.1 \wedge 0) \vee (1 \wedge 0) \vee (0.1 \wedge 0.7) \\
&= 0.1.
\end{aligned}$$

We end this section by making a close link between computing with values and computing with words.

*Theorem 1:* Suppose that $\tilde{M} = (Q, \tilde{\Sigma}, \tilde{\delta}, q_0, F)$ is an FACW and $\tilde{M}^{\downarrow} = (Q, \Sigma, \delta^{\downarrow}, q_0, F)$ is the retraction of $\tilde{M}$. Then for any $w = a_1 \cdots a_n \in \Sigma^*$, we have that

$$L(\tilde{M}^{\downarrow})(w) = \vee_{A_1,\ldots,A_n \in \tilde{\Sigma}} \big[L_w(\tilde{M})(A_1 \cdots A_n) \wedge A_1(a_1) \wedge \cdots \wedge A_n(a_n)\big].$$

*Proof:* See Appendix I. ∎

The above theorem may be seen as a retraction principle from computing with words to computing with values. The meaning of this theorem is that computing with values can be implemented by computing with words; and thus it gives us a way of dealing with crisp inputs on an FACW. Observe that the number of computations for implementing computing with values by computing with words increases exponentially as the length of the input string.

## IV. GENERALIZED EXTENSIONS: COMPUTING WITH ALL WORDS

Having finished the transformation from computing with words to computing with values in the preceding section, we turn our attention to another transformation which makes an FACW more robust in the sense that it can deal with some inputs not in its input alphabet.

More explicitly, let us consider the following problem. Suppose that there is an FACW $\tilde{M} = (Q, \tilde{\Sigma}, \tilde{\delta}, q_0, F)$. Note that the input alphabet $\tilde{\Sigma}$ comprises only some (not necessarily all) words (i.e., fuzzy subsets of $\Sigma$). This means that $\tilde{M}$ cannot accept a fuzzy subset $A'$, which is in $\mathcal{F}(\Sigma)$ but not in $\tilde{\Sigma}$, as an input, although $A'$ may be very similar to $A$. For instance, $A = small$ and $A' = almost\ small$. Since the fuzzy sets in $\tilde{\Sigma}$ are mathematical expressions of linguistic terms that are usually selected by an expert and are always somewhat imprecise and vague, it is not reasonable to discriminate among similar inputs. Thus, if $\tilde{\delta}(q, A)$ is defined for some $(q, A) \in Q \times \tilde{\Sigma}$, we hope that $\tilde{\delta}(q, A')$ would be defined whenever $A'$ is similar to $A$. To this end, we will extend $\tilde{\delta}$ to a fuzzy transition function $\delta^{\uparrow}$ from $Q \times \mathcal{F}(\Sigma)$ to $\mathcal{F}(Q)$. As a result, we will obtain an FACAW $\tilde{M}^{\uparrow} = (Q, \mathcal{F}(\Sigma), \delta^{\uparrow}, q_0, F)$ which can accept more words than in $\tilde{\Sigma}$ as inputs.

Like the retractions, we define $\delta^{\uparrow}$ again by utilizing the methodology of fuzzy control. Recall that the fuzzification of FLS introduced in the previous section can merely cope with crisp inputs. In order to handle fuzzy inputs, we have to seek other fuzzification techniques. In [7], Foulloy and Galichet provided a means of introducing fuzzy inputs in Takagi–Sugeno–Kang type FLS. In a similar manner, we can develop a fuzzification for dealing with fuzzy inputs in our context.

For later need, let us first review some basic concepts in [7]. Let $\mathcal{L}$ be a set of linguistic terms and $X$ be a crisp set. Suppose that $R$ is a fuzzy relation defined on $\mathcal{L} \times X$, that is, $R \in \mathcal{F}(\mathcal{L} \times X)$. Now to any linguistic term $l \in \mathcal{L}$, we associate a fuzzy subset $M_l \in \mathcal{F}(X)$ given by the membership function

$$M_l(x) = R(l, x)$$

for any $x \in X$. The fuzzy subset $M_l$ is called the *fuzzy meaning* of $l$. Obviously, the fuzzy meaning is just another representation of a fuzzy relation. In the same way, to any $x \in X$, we associate a fuzzy subset $D_x \in \mathcal{F}(\mathcal{L})$ given by the membership function

$$D_x(l) = R(l, x)$$

for any $l \in \mathcal{L}$. The fuzzy subset $D_x$ is called the *fuzzy description* of $x$. The fuzzy description is a simple means of describing an element of $X$ in words. This notion can be generalized to fuzzy subsets. For any $A \in \mathcal{F}(X)$, the *(upper) fuzzy description* of $A$ is a fuzzy subset $D_A \in \mathcal{F}(\mathcal{L})$ given by the membership function

$$D_A(l) = \text{height}(M_l \cap A)$$

for any $l \in \mathcal{L}$, which represents the possibility of a fuzzy event characterized by the fuzzy meaning $M_l$ of the linguistic terms $l \in \mathcal{L}$, considering the fuzzy input $A$ as a possibility distribution. As interpreted in [7], the fuzzy description of a fuzzy subset gives the linguistic terms which possibly describe an input.

We now use the fuzzy description to determine the matching degree between a fuzzy input and the premise of a fuzzy IF-THEN rule. Assume that the current state of $\tilde{M}^\uparrow$ is $q$ and the input of $\tilde{M}^\uparrow$ is $A' \in \mathcal{F}(\Sigma)$, and let $\mathrm{R}_p^A$ be a rule in $\mathfrak{R}$. For the first premise of $\mathrm{R}_p^A$, we may take $\mathcal{L} = \mathcal{F}(Q)$ and $X = Q$. Let $R$ be the fuzzy relation that maps $(\tilde{p}', q')$ to $\tilde{p}'(q')$ for any $(\tilde{p}', q') \in \mathcal{F}(Q) \times Q$. Since the current state $q$ of $\tilde{M}^\uparrow$ is crisp, we may take $D_q(1/p)$ as the fuzzy description of $q$ relative to the rule $\mathrm{R}_p^A$. By definition,

$$D_q(\frac{1}{p}) = R(\frac{1}{p}, q) = \frac{1}{p}(q).$$

For the second premise, we may take $\mathcal{L} = \tilde{\Sigma}$ and $X = \Sigma$. Let $R$ be the fuzzy relation sending $(B, a)$ to $B(a)$ for any $(B, a) \in \tilde{\Sigma} \times \Sigma$. Consequently, the fuzzy meaning of $A$ is given by

$$M_A(a) = R(A, a) = A(a)$$

for any $a \in \Sigma$. We thus obtain that the fuzzy description of $A'$ relative to the rule $\mathrm{R}_p^A$ is

$$\begin{aligned} D_{A'}(A) &= \mathrm{height}(M_A \cap A') \\ &= \vee_{a \in \Sigma}[M_A(a) \wedge A'(a)] \\ &= \vee_{a \in \Sigma}[A(a) \wedge A'(a)]. \end{aligned}$$

Further, for any fuzzy IF-THEN rule $\mathrm{R}_p^A$ we define a fuzzy set $\mathbb{M}_p^A \in \mathcal{F}(Q \times \mathcal{F}(\Sigma))$ with the membership function given by

$$\mathbb{M}_p^A(q, A') = D_q(\frac{1}{p}) \wedge D_{A'}(A).$$

Similar to the FLS with crisp inputs, $\mathbb{M}_p^A(q, A')$ can be viewed as the matching degree between the current situation of $\tilde{M}^\uparrow$ and the premise of $\mathrm{R}_p^A$. By a straightforward calculation, we have that

$$\mathbb{M}_p^A(q, A') = \begin{cases} \vee_{a \in \Sigma}[A(a) \wedge A'(a)], & \text{if } q = p \\ 0, & \text{otherwise.} \end{cases}$$

After finishing the step of matching, let us address the inference step. In fact, the inference step of FLS introduced in the last section remains applicable. Correspondingly, the *implied fuzzy set* for the rule $\mathrm{R}_p^A$, denoted $\mathbb{N}_p^A$, is defined by the membership function

$$\mathbb{N}_p^A(q') = \mathbb{M}_p^A(q, A') \wedge \tilde{\delta}(p, A)(q')$$

for any $q' \in Q$. The implied fuzzy set $\mathbb{N}_p^A$ specifies the certainty level that the next state should be $q'$ when only the rule $\mathrm{R}_p^A$ is considered. The *overall implied fuzzy set*, denoted $\bar{\mathbb{N}}$, is given by the membership function

$$\bar{\mathbb{N}}(q') = \vee_{(p, A) \in Q \times \tilde{\Sigma}} \mathbb{N}_p^A(q')$$

for any $q' \in Q$. The membership value $\bar{\mathbb{N}}(q')$ is a result of considering all the rules in the rule-base at the same time, and it gives the maximal possibility degree of $q'$ as the next state when the current state is $q$ and the input is $A'$.

In the same manner as for the retraction, we define $\delta^\uparrow(q, A') = \bar{\mathbb{N}}$. By the previous definitions, we have that

$$\begin{aligned} \delta^\uparrow(q, A')(q') &= \bar{\mathbb{N}}(q') \\ &= \vee_{(p, A) \in Q \times \tilde{\Sigma}} \mathbb{N}_p^A(q') \\ &= \vee_{(p, A) \in Q \times \tilde{\Sigma}} [\mathbb{M}_p^A(q, A') \wedge \tilde{\delta}(p, A)(q')] \\ &= \vee_{(q, A) \in \{q\} \times \tilde{\Sigma}} [\mathbb{M}_q^A(q, A') \wedge \tilde{\delta}(q, A)(q')] \\ &= \vee_{A \in \tilde{\Sigma}} \{ [\vee_{a \in \Sigma} [A(a) \wedge A'(a)]] \\ &\quad \wedge \tilde{\delta}(q, A)(q') \} \\ &= \vee_{A \in \tilde{\Sigma}} \vee_{a \in \Sigma} [A(a) \wedge A'(a) \wedge \tilde{\delta}(q, A)(q')], \end{aligned}$$

i.e., $\delta^\uparrow(q, A')(q') = \vee_{A \in \tilde{\Sigma}} \vee_{a \in \Sigma} [A(a) \wedge A'(a) \wedge \tilde{\delta}(q, A)(q')]$ for any $q' \in Q$.

Finally, we have the following definition.

*Definition 6:* Let $\tilde{M} = (Q, \tilde{\Sigma}, \tilde{\delta}, q_0, F)$ be an FACW. The *generalized extension* of $\tilde{M}$ is an FACAW $\tilde{M}^\uparrow = (Q, \mathcal{F}(\Sigma), \delta^\uparrow, q_0, F)$, where the components $Q, q_0, F$ are the same as those of $\tilde{M}$, $\mathcal{F}(\Sigma)$ consists of all fuzzy subsets of the underlying input alphabet of $\tilde{M}$, and $\delta^\uparrow$ is a mapping from $Q \times \mathcal{F}(\Sigma)$ to $\mathcal{F}(Q)$ defined by

$$\delta^\uparrow(q, A')(q') = \vee_{A \in \tilde{\Sigma}} \vee_{a \in \Sigma} [A(a) \wedge A'(a) \wedge \tilde{\delta}(q, A)(q')] \quad (1)$$

for any $(q, A') \in Q \times \mathcal{F}(\Sigma)$ and $q' \in Q$.

As we see from the above definition, the generalized extension $\tilde{M}^\uparrow$ of $\tilde{M}$ can deal with all words over the underlying input alphabet of $\tilde{M}$ as inputs. We thus consider $\tilde{M}^\uparrow$ as a device for computing with all words and refer to "$\uparrow$" as the operation of obtaining the generalized extension. The differences between $\tilde{M}^\uparrow$ and $\tilde{M}$ are visible, as the following example shows.

*Example 3:* Consider again the FACW $\tilde{M} = (Q, \tilde{\Sigma}, \tilde{\delta}, q_0, F)$ given in Example 1. By Definition 6, we get that the generalized extension of $\tilde{M}$ is $\tilde{M}^\uparrow = (Q, \mathcal{F}(\Sigma), \delta^\uparrow, q_0, F)$, where $\delta^\uparrow$ is given by (1) in Definition 6. For instance, let $S'$ be $almost\ small$ defined by the membership function $S'(x) = [S(x)]^{\frac{1}{2}}$ for any $x \in \Sigma = \{1, 2, 3, 4, 5\}$, where $S = small = 1/1 + 0.5/2 + 0.1/3$. By a simple calculation, we see that $S' = 1/1 + 0.7071/2 + 0.3162/3 \notin \tilde{\Sigma}$, which means that $S'$ is not an admissible input of $\tilde{M}$. But $S' \in \mathcal{F}(\Sigma)$, so $S'$ is an admissible input of $\tilde{M}^\uparrow$. If we input the word $S'$ at $q_0$ of $\tilde{M}^\uparrow$, then we obtain that

$$\begin{aligned} \delta^\uparrow(q_0, S')(q_0) &= \vee_{A \in \tilde{\Sigma}} \vee_{a \in \Sigma} [A(a) \wedge S'(a) \wedge \tilde{\delta}(q_0, A)(q_0)] \\ &= 1, \\ \delta^\uparrow(q_0, S')(q_1) &= \vee_{A \in \tilde{\Sigma}} \vee_{a \in \Sigma} [A(a) \wedge S'(a) \wedge \tilde{\delta}(q_0, A)(q_1)] \\ &= 0.3162, \\ \delta^\uparrow(q_0, S')(q_2) &= \vee_{A \in \tilde{\Sigma}} \vee_{a \in \Sigma} [A(a) \wedge S'(a) \wedge \tilde{\delta}(q_0, A)(q_2)] \\ &= 0.1, \end{aligned}$$

namely, $\delta^\uparrow(q_0, S') = 1/q_0 + 0.3162/q_1 + 0.1/q_2$.

Analogous to Theorem 1, we can also establish a close link between computing with some special words (i.e., those in $\tilde{\Sigma}$) and computing with all words.

*Theorem 2:* Suppose that $\tilde{M} = (Q, \tilde{\Sigma}, \tilde{\delta}, q_0, F)$ is an FACW and $\tilde{M}^\uparrow = (Q, \mathcal{F}(\Sigma), \delta^\uparrow, q_0, F)$ is the generalized extension of $\tilde{M}^\uparrow$. Then for any $W = A'_1 \cdots A'_n \in \mathcal{F}(\Sigma)^*$, we have that

$$L_w(\tilde{M}^\uparrow)(W) = \vee_{A_1,\ldots,A_n \in \tilde{\Sigma}} \vee_{a_1,\ldots,a_n \in \Sigma} \big[L_w(\tilde{M})(A_1 \cdots A_n) \\ \wedge A_1(a_1) \wedge \cdots \wedge A_n(a_n) \wedge A'_1(a_1) \wedge \cdots \wedge A'_n(a_n)\big].$$

*Proof:* See Appendix I. ∎

Theorem 2 may be seen as a generalized extension principle from computing with special words to computing with all words. The meaning of this theorem is that computing with all words can be implemented by computing with special words; and thus it gives us a way to deal with arbitrary fuzzy inputs on an FACW. It is clear that the number of computations for implementing computing with all words by computing with words increases exponentially as the length of the input string.

## V. Relationships among Retractions, Extensions, and Generalized Extensions

Up to now, we have seen three kinds of transformations among FACVs, FACWs, and FACAWs, that is, the extensions in Definition 3, the retractions, and the generalized extensions. In fact, they are related; some of the relationships are shown in this section.

We first show that the extension given in Definition 3 is a special case of the generalized extension introduced in the last section. To see this, we only need to regard an FACV as an FACW by identifying an input with its singleton fuzzification. Given an FACV $M' = (Q', \Sigma', \delta', q'_0, F')$, it is clear that we can identify $M'$ with an FACW $\tilde{M}' = (Q', \tilde{\Sigma}', \tilde{\delta}', q'_0, F')$, where two different components are

$\tilde{\Sigma}' = \{\tilde{a} : \tilde{a}$ is the singleton fuzzification of $a \in \Sigma\}$ and
$\tilde{\delta}'(p, \tilde{a}) = \delta'(p, a)$ for any $p \in Q'$ and $\tilde{a} \in \tilde{\Sigma}'$.

By definition, it is easy to verify that $\tilde{M}'^\downarrow = M'$.

Further, we have the following.

*Proposition 1:* Let $M' = (Q', \Sigma', \delta', q'_0, F')$ be an FACV. Then $\hat{M}' = (Q', \mathcal{F}(\Sigma'), \hat{\delta}', q'_0, F')$ given by Definition 3 is the same as the generalized extension $\tilde{M}'^\uparrow = (Q', \mathcal{F}(\Sigma'), \delta'^\uparrow, q'_0, F')$.

*Proof:* It is sufficient to show that $\hat{\delta}'(p, A)(q) = \delta'^\uparrow(p, A)(q)$ for any $p, q \in Q'$ and $A \in \mathcal{F}(\Sigma')$. By Definition 3, we see that $\hat{\delta}'(p, A)(q) = \vee_{a \in \Sigma'}[A(a) \wedge \delta'(p, a)(q)]$. On the other hand, it follows from Definition 6 that

$$\begin{aligned}\delta'^\uparrow(p, A)(q) &= \vee_{\tilde{a} \in \tilde{\Sigma}'} \vee_{b \in \Sigma'} [\tilde{a}(b) \wedge A(b) \wedge \tilde{\delta}'(p, \tilde{a})(q)] \\ &= \vee_{\tilde{a} \in \tilde{\Sigma}'} \vee_{b \in \Sigma'} [\tilde{a}(b) \wedge A(b) \wedge \delta'(p, a)(q)] \\ &= \vee_{\tilde{a} \in \tilde{\Sigma}'} [\tilde{a}(a) \wedge A(a) \wedge \delta'(p, a)(q)] \\ &= \vee_{a \in \Sigma'} [A(a) \wedge \delta'(p, a)(q)].\end{aligned}$$

Hence, $\hat{\delta}'(p, A)(q) = \delta'^\uparrow(p, A)(q)$, as desired. ∎

Based on Proposition 1, we view the extension in Definition 3 as a generalized extension hereafter. As we see from Fig. 1, there are two approaches from computing with words to computing with all words: One is the generalized extension $(b)$; the other is the composition of processes $(a)$ and $(c)$. The next proposition shows that the two approaches yield the same result.

*Proposition 2:* Let $\tilde{M} = (Q, \tilde{\Sigma}, \tilde{\delta}, q_0, F)$ be an FACW. Then $(\tilde{M}^\downarrow)^\uparrow = \tilde{M}^\uparrow$.

*Proof:* By definition, $\tilde{M}^\uparrow = (Q, \mathcal{F}(\Sigma), \delta^\uparrow, q_0, F)$ with $\delta^\uparrow(p, A)(q) = \vee_{A' \in \tilde{\Sigma}} \vee_{a \in \Sigma}[A'(a) \wedge A(a) \wedge \tilde{\delta}(p, A')(q)]$ for any $p, q \in Q$ and $A \in \mathcal{F}(\Sigma)$. In contrast, $\tilde{M}^\downarrow = (Q, \Sigma, \delta^\downarrow, q_0, F)$, where $\delta^\downarrow(p, a)(q) = \vee_{A' \in \tilde{\Sigma}}[A'(a) \wedge \tilde{\delta}(p, A')(q)]$ for any $p, q \in Q$ and $a \in \Sigma$. Consequently, the generalized extension of $\tilde{M}^\downarrow$ is $(\tilde{M}^\downarrow)^\uparrow = (Q, \mathcal{F}(\Sigma), (\delta^\downarrow)^\uparrow, q_0, F)$. By Proposition 1 and definition, we see that

$$\begin{aligned}(\delta^\downarrow)^\uparrow(p, A)(q) &= \vee_{a \in \Sigma}[A(a) \wedge \delta^\downarrow(p, a)(q)] \\ &= \vee_{a \in \Sigma}\{A(a) \wedge [\vee_{A' \in \tilde{\Sigma}}[A'(a) \\ &\qquad \wedge \tilde{\delta}(p, A')(q)]]\} \\ &= \vee_{a \in \Sigma}\{\vee_{A' \in \tilde{\Sigma}}[A'(a) \wedge A(a) \wedge \tilde{\delta}(p, A')(q)]\} \\ &= \vee_{A' \in \tilde{\Sigma}} \vee_{a \in \Sigma}[A'(a) \wedge A(a) \wedge \tilde{\delta}(p, A')(q)] \\ &= \delta^\uparrow(p, A)(q)\end{aligned}$$

for any $p, q \in Q$ and $A \in \mathcal{F}(\Sigma)$. That is, $(\delta^\downarrow)^\uparrow = \delta^\uparrow$, and thus $(\tilde{M}^\downarrow)^\uparrow = \tilde{M}^\uparrow$, finishing the proof. ∎

A careful reader may find that the generalized extension from $\tilde{M}$ to $\tilde{M}^\uparrow$ is generally not an extension in the strictly mathematical sense, that is, $\delta^\uparrow(p, A)$ is not necessarily equal to $\tilde{\delta}(p, A)$ even for $A \in \tilde{\Sigma}$. To see this, let us revisit Example 3. Keep all notations in Example 3. We see that $\tilde{\delta}(q_0, S) = 1/q_0$, whereas by a computation analogous to that of $\delta^\uparrow(q_0, S')$, we find that $\delta^\uparrow(q_0, S) = 1/q_0 + 0.2/q_1 + 0.1/q_2$. Further, we get that $L_w(\tilde{M})(S) = 0.1$ and $L_w(\tilde{M}^\uparrow)(S) = 0.2$; they are not equal. This motivates us to consider the consistency of the generalized extension.

From the viewpoint of computing, we are interested in the following property. Let $\tilde{M} = (Q, \tilde{\Sigma}, \tilde{\delta}, q_0, F)$ be an FACW with the generalized extension $\tilde{M}^\uparrow = (Q, \mathcal{F}(\Sigma), \delta^\uparrow, q_0, F)$. If $L_w(\tilde{M}^\uparrow)(W) = L_w(\tilde{M})(W)$ holds for all $W \in \tilde{\Sigma}^*$, then the generalized extension from computing with words to computing with all words is called *consistent*. The consistency implies that for any input $W \in \tilde{\Sigma}^*$, using $\tilde{M}^\uparrow$ as a computing device is the same as using $\tilde{M}$. If each word in $\tilde{\Sigma}$ degenerates into a singleton, then it is not hard to check that the generalized extension is consistent. In general, not all generalized extensions are consistent, as we have seen above.

The direct reason for the failure of consistency is that $\delta^\uparrow(p, A) \neq \tilde{\delta}(p, A)$ for some $p \in Q$ and $A \in \tilde{\Sigma}$. The appearance of such an inequality is not surprising if we have noticed that the calculation of $\delta^\uparrow(p, A)$ depends on all $A' \in \tilde{\Sigma}$ and $\tilde{\delta}(p, A')$, while the words $A' \in \tilde{\Sigma}$ may be intersecting each other. Clearly, if the calculation of $\delta^\uparrow(p, A)$ is not disturbed by those $A' \in \tilde{\Sigma} \setminus \{A\}$ and $\tilde{\delta}(p, A')$, then the generalized extension must be consistent. From this point of view, the consistency measures the independence of information afforded by the words and fuzzy transition function of $\tilde{M}$. So we introduce the following definition.

*Definition 7:* Let $\tilde{M} = (Q, \tilde{\Sigma}, \tilde{\delta}, q_0, F)$ be an FACW. The *independence degree* of $\tilde{\Sigma}$ and $\tilde{\delta}$, denoted $d(\tilde{M})$, is defined by

$$d(\tilde{M}) = \sup_{W \in \tilde{\Sigma}^*} |L_w(\tilde{M}^\uparrow)(W) - L_w(\tilde{M})(W)|,$$

where $\tilde{M}^\uparrow$ is the generalized extension of $\tilde{M}$.

By definition, the generalized extension from $\tilde{M}$ to $\tilde{M}^\uparrow$ is consistent if and only if the independence degree equals 0. If $\tilde{M}$ is specified by an expert, then $d(\tilde{M}) > 0$ is easily understandable, since the information given by experts is usually not completely independent. For a given $\tilde{M}$ with $d(\tilde{M}) > 0$, one can utilize the idea of [29] and [26] to slightly modify the fuzzy transition function $\tilde{\delta}$ and then gives a new FACW $\tilde{M}'$ such that $d(\tilde{M}') = 0$. The detail of revising $\tilde{\delta}$ is beyond the scope of this paper, so we do not discuss it here.

Clearly, a sufficient condition for the independence degree to be zero is that $\delta^\uparrow(p, A) = \tilde{\delta}(p, A)$ for any $p \in Q$ and $A \in \tilde{\Sigma}$, namely, $\delta^\uparrow|_{Q \times \tilde{\Sigma}} = \tilde{\delta}$, where the notation $\varphi|_{X'}$ means that we are restricting the mapping $\varphi$ defined on $X$ to the smaller domain $X'$. The equality $\delta^\uparrow|_{Q \times \tilde{\Sigma}} = \tilde{\delta}$ implies that the information is preserved when extending $\tilde{M}$ to $\tilde{M}^\uparrow$. We end this section with a characterization of the equality.

*Proposition 3:* Let $\tilde{M} = (Q, \tilde{\Sigma}, \tilde{\delta}, q_0, F)$ be an FACW with the generalized extension $\tilde{M}^\uparrow = (Q, \mathcal{F}(\Sigma), \delta^\uparrow, q_0, F)$. Then $\delta^\uparrow|_{Q \times \tilde{\Sigma}} = \tilde{\delta}$ if and only if for any $p, q \in Q$ and $A \in \tilde{\Sigma}$, the following hold:
1) there exists $a \in \Sigma$ such that $A(a) \geq \tilde{\delta}(p, A)(q)$;
2) for any $a \in \{a \in \Sigma : A(a) > \tilde{\delta}(p, A)(q)\}$ and $A' \in \tilde{\Sigma} \setminus \{A\}$, either $A'(a) \leq \tilde{\delta}(p, A)(q)$ or $\tilde{\delta}(p, A')(q) \leq \tilde{\delta}(p, A)(q)$.

*Proof:* By definition, we see that $\delta^\uparrow|_{Q \times \tilde{\Sigma}} = \tilde{\delta}$ if and only if
$$\vee_{A' \in \tilde{\Sigma}} \vee_{a \in \Sigma} [A'(a) \wedge A(a) \wedge \tilde{\delta}(p, A')(q)] = \tilde{\delta}(p, A)(q) \quad (2)$$
for any $p, q \in Q$ and $A \in \tilde{\Sigma}$.

Let us first show the 'only if' part. For 1), suppose, by contradiction, that $A(a) < \tilde{\delta}(p, A)(q)$ for all $a \in \Sigma$. Then it follows that
$$\vee_{A' \in \tilde{\Sigma}} \vee_{a \in \Sigma} [A'(a) \wedge A(a) \wedge \tilde{\delta}(p, A')(q)] \leq A(a) < \tilde{\delta}(p, A)(q),$$
which contradicts (2). Hence, 1) holds. The condition 2) can also be easily proven by contradiction.

We now prove the 'if' part. Using 1), we get that
$$\vee_{A' \in \tilde{\Sigma}} \vee_{a \in \Sigma} [A'(a) \wedge A(a) \wedge \tilde{\delta}(p, A')(q)]$$
$$\geq \vee_{a \in \Sigma} [A(a) \wedge A(a) \wedge \tilde{\delta}(p, A)(q)]$$
$$\geq \tilde{\delta}(p, A)(q).$$

On the other hand, it follows directly from 2) that $\vee_{A' \in \tilde{\Sigma}} \vee_{a \in \Sigma} [A'(a) \wedge A(a) \wedge \tilde{\delta}(p, A')(q)] \leq \tilde{\delta}(p, A)(q)$. Therefore, (2) is true, and thus $\delta^\uparrow|_{Q \times \tilde{\Sigma}} = \tilde{\delta}$, finishing the proof of the proposition. ∎

## VI. SOME ALGEBRAIC PROPERTIES OF RETRACTIONS AND GENERALIZED EXTENSIONS

In this section, we look at the preservation properties of retractions and generalized extensions under the use of product operator and homomorphism.

As in classical computing, we can build the overall model of computing with words by building models of individual components first and then composing them by product. The product operation models a form of joint behavior of a set of FACWs and we can think of it as one type of systems resulting from the interconnection of system components. Let us recall the general definition (cf. [1], [15], [19] for some relevant notions).

Let $M_i = (Q_i, \Sigma, \delta_i, q_{0i}, F_i)$ be a fuzzy automaton, where $i = 1, 2$. The *product* of $M_1$ and $M_2$ is a fuzzy automaton
$$M_1 \times M_2 = (Q_1 \times Q_2, \Sigma, \delta_1 \wedge \delta_2, (q_{01}, q_{02}), F),$$
where
$$\delta_1 \wedge \delta_2((p_1, q_1), \sigma)(p_2, q_2) = \delta_1(p_1, a)(p_2) \wedge \delta_2(q_1, a)(q_2)$$
for all $(p_i, q_i) \in Q_1 \times Q_2$ and $\sigma \in \Sigma$, and $F$ is a fuzzy subset of $Q_1 \times Q_2$ with the membership function $F(p, q) = F_1(p) \wedge F_2(q)$ for any $(p, q) \in Q_1 \times Q_2$. It is easy to verify that $L(M_1 \times M_2) = L(M_1) \cap L(M_2)$.

The next proposition shows that the language (resp. word language) accepted by the retraction (resp. generalized extension) of the product is bounded by the language (resp. word language) accepted by the individual retraction of each component.

*Proposition 4:* Let $\tilde{M}_i = (Q_i, \tilde{\Sigma}, \tilde{\delta}_i, q_{0i}, F_i)$ be an FACW with the retraction $\tilde{M}_i^\downarrow$ and the generalized extension $\tilde{M}_i^\uparrow$, $i = 1, 2$. Then $L((\tilde{M}_1 \times \tilde{M}_2)^\downarrow) \subseteq L(\tilde{M}_1^\downarrow) \cap L(\tilde{M}_2^\downarrow)$ and $L_w((\tilde{M}_1 \times \tilde{M}_2)^\uparrow) \subseteq L_w(\tilde{M}_1^\uparrow) \cap L_w(\tilde{M}_2^\uparrow)$.

*Proof:* We only prove the first inclusion; the second one can be proved similarly. Assume that $\tilde{M}_1^\downarrow = (Q_1, \Sigma, \delta_1^\downarrow, q_{01}, F_1)$, $\tilde{M}_2^\downarrow = (Q_2, \Sigma, \delta_2^\downarrow, q_{02}, F_2)$, $\tilde{M}_1^\downarrow \times \tilde{M}_2^\downarrow = (Q_1 \times Q_2, \Sigma, \delta_1^\downarrow \wedge \delta_2^\downarrow, (q_{01}, q_{02}), F')$, and $(\tilde{M}_1 \times \tilde{M}_2)^\downarrow = (Q_1 \times Q_2, \Sigma, \tilde{\delta}^\downarrow, (q_{01}, q_{02}), F'')$. By definition, we see that $F' = F''$, and for any $(p_1, q_1), (p_2, q_2) \in Q_1 \times Q_2$ and $a \in \Sigma$,
$$\delta_1^\downarrow \wedge \delta_2^\downarrow((p_1, q_1), a)(p_2, q_2) = \delta_1^\downarrow(p_1, a)(p_2) \wedge \delta_2^\downarrow(q_1, a)(q_2)$$
and
$$\tilde{\delta}^\downarrow((p_1, q_1), a)(p_2, q_2)$$
$$= \vee_{A \in \tilde{\Sigma}}[A(a) \wedge (\tilde{\delta}_1 \wedge \tilde{\delta}_2)((p_1, q_1), A)(p_2, q_2)]$$
$$= \vee_{A \in \tilde{\Sigma}}[A(a) \wedge \tilde{\delta}_1(p_1, A)(p_2) \wedge \tilde{\delta}_2(q_1, A)(q_2)].$$

Recall that $L(\tilde{M}_1^\downarrow \times \tilde{M}_2^\downarrow) = L(\tilde{M}_1^\downarrow) \cap L(\tilde{M}_2^\downarrow)$. Thereby, we need only to prove that $L((\tilde{M}_1 \times \tilde{M}_2)^\downarrow) \subseteq L(\tilde{M}_1^\downarrow \times \tilde{M}_2^\downarrow)$. It is enough to show that $\tilde{\delta}^\downarrow((p_1, q_1), a)(p_2, q_2) \leq \delta_1^\downarrow \wedge \delta_2^\downarrow((p_1, q_1), a)(p_2, q_2)$ for any $(p_1, q_1), (p_2, q_2) \in Q_1 \times Q_2$ and $a \in \Sigma$. In fact, by the previous argument we have that
$$\tilde{\delta}^\downarrow((p_1, q_1), a)(p_2, q_2)$$
$$= \vee_{A \in \tilde{\Sigma}}[A(a) \wedge \tilde{\delta}_1(p_1, A)(p_2) \wedge \tilde{\delta}_2(q_1, A)(q_2)]$$
$$\leq \{\vee_{A \in \tilde{\Sigma}}[A(a) \wedge \tilde{\delta}_1(p_1, A)(p_2)]\}$$
$$\qquad \wedge \{\vee_{A \in \tilde{\Sigma}}[A(a) \wedge \tilde{\delta}_2(q_1, A)(q_2)]\} \quad (3)$$
$$= \delta_1^\downarrow(p_1, a)(p_2) \wedge \delta_2^\downarrow(q_1, a)(q_2)$$
$$= \delta_1^\downarrow \wedge \delta_2^\downarrow((p_1, q_1), a)(p_2, q_2),$$

as desired. This finishes the proof of the proposition. ∎

It is easy to observe that the inequality (3) appearing in the proof of Proposition 4 can be strict, so the inclusions in Proposition 4 can also be strict.

Inputting the same word at a state means different next state distributions to different experts. The concept of homomorphism can relate these distributions. We end this section by discussing the preservation of the generalized extension under a homomorphism.

Given two fuzzy automata $M_i = (Q_i, \Sigma, \delta_i, q_{0i}, F_i)$, $i = 1, 2$, we say that $M_1$ is a *subfuzzy automaton* of $M_2$, written $M_1 \leq M_2$, if $Q_1 \subseteq Q_2$, $q_{01} = q_{02}$, $F_1 \subseteq F_2$, and $\delta_1 = \delta_2|_{Q_1 \times \Sigma}$.

*Definition 8:* Let $M_1 = (Q_1, \Sigma, \delta_1, q_{01}, F_1)$ and $M_2 = (Q_2, \Sigma, \delta_2, q_{02}, F_2)$ be two fuzzy automata. A mapping $f : Q_1 \longrightarrow Q_2$ is called a *homomorphism* from $M_1$ to $M_2$ if the following hold:

1) $f(q_{01}) = q_{02}$.
2) $\delta_2(f(p), \sigma)(f(q)) = \vee\{\delta_1(p, \sigma)(r) : r \in Q_1, f(r) = f(q)\}$ for any $p, q \in Q_1$ and $\sigma \in \Sigma$.
3) $F_1(q) \leq F_2(f(q))$ for any $q \in Q_1$.

The *homomorphism image* of $M_1$ under a homomorphism $f$, denoted $f(M_1)$, is defined as $(f(Q_1), \Sigma, \delta_2|_{f(Q_1) \times \Sigma}, q_{02}, F_2|_{f(Q_1)})$; it is clear that $f(M_1) \leq M_2$. In particular, if $M_1 \leq M_2$, the embedding mapping $i : Q_1 \hookrightarrow Q_2$ gives rise to a homomorphism; the homomorphism image of $M_1$ under $i$ is identical with itself. Further, we make the following observation.

*Lemma 1:* Let $M_1 = (Q_1, \Sigma, \delta_1, q_{01}, F_1)$ and $M_2 = (Q_2, \Sigma, \delta_2, q_{02}, F_2)$ be two fuzzy automata. If there is a homomorphism from $M_1$ to $M_2$, then $L(M_1) \subseteq L(M_2)$.

*Proof:* It follows readily from Definitions 2 and 8. ∎

As a result, $L(M_1) = L(M_2)$ whenever there exists a pair of homomorphisms $f : M_1 \longrightarrow M_2$ and $g : M_2 \longrightarrow M_1$. This implies that we may compare the computing power of two fuzzy automata by constructing a homomorphism between them.

If two FACWs are related by a homomorphism $f$, then so are their retractions (resp. generalized extensions); moreover, the homomorphism $f$ preserves retractions (resp. generalized extensions). More formally, we have the following proposition.

*Proposition 5:* Let $\tilde{M}_i = (Q_i, \tilde{\Sigma}, \tilde{\delta}_i, q_{0i}, F_i)$ be an FACW with the retraction $\tilde{M}_i^\downarrow$ and the generalized extension $\tilde{M}_i^\uparrow$, $i = 1, 2$. If $f$ is a homomorphism from $\tilde{M}_1$ to $\tilde{M}_2$, then:

1) $f$ gives a homomorphism from $\tilde{M}_1^\downarrow$ to $\tilde{M}_2^\downarrow$, and, moreover, $f(\tilde{M}_1)^\downarrow = f(\tilde{M}_1^\downarrow)$.
2) $f$ gives a homomorphism from $\tilde{M}_1^\uparrow$ to $\tilde{M}_2^\uparrow$, and, moreover, $f(\tilde{M}_1)^\uparrow = f(\tilde{M}_1^\uparrow)$.

*Proof:* We only prove the assertion 1), and 2) can be proved analogously. Since $f$ is a homomorphism from $\tilde{M}_1$ to $\tilde{M}_2$, $f$ is a mapping from $Q_1$ to $Q_2$ that satisfies

- $f(q_{01}) = q_{02}$;
- $\tilde{\delta}_2(f(p), A)(f(q)) = \vee\{\tilde{\delta}_1(p, A)(r) : r \in Q_1, f(r) = f(q)\}$ for any $p, q \in Q_1$ and $A \in \tilde{\Sigma}$; and
- $F_1(q) \leq F_2(f(q))$ for any $q \in Q_1$.

To prove that $f$ is a homomorphism from $\tilde{M}_1^\downarrow = (Q_1, \Sigma, \delta_1^\downarrow, q_{01}, F_1)$ to $\tilde{M}_2^\downarrow = (Q_2, \Sigma, \delta_2^\downarrow, q_{02}, F_2)$, it is enough to show that $\delta_2^\downarrow(f(p), a)(f(q)) = \vee\{\delta_1^\downarrow(p, a)(r) : r \in Q_1, f(r) = f(q)\}$ for any $p, q \in Q_1$ and $a \in \Sigma$. In fact,

it follows from definition that

$$\begin{aligned}
\delta_2^\downarrow(f(p), a)(f(q)) &= \vee_{A \in \tilde{\Sigma}}[A(a) \wedge \tilde{\delta}_2(f(p), A)(f(q))] \\
&= \vee_{A \in \tilde{\Sigma}}\{A(a) \wedge [\vee_{\substack{r \in Q_1 \\ f(r) = f(q)}} \tilde{\delta}_1(p, A)(r)]\} \\
&= \vee_{A \in \tilde{\Sigma}} \vee_{\substack{r \in Q_1 \\ f(r) = f(q)}} [A(a) \wedge \tilde{\delta}_1(p, A)(r)] \\
&= \vee_{\substack{r \in Q_1 \\ f(r) = f(q)}} \vee_{A \in \tilde{\Sigma}} [A(a) \wedge \tilde{\delta}_1(p, A)(r)] \\
&= \vee_{\substack{r \in Q_1 \\ f(r) = f(q)}} \delta_1^\downarrow(p, a)(r) \\
&= \vee\{\delta_1^\downarrow(p, a)(r) : r \in Q_1, f(r) = f(q)\}
\end{aligned}$$

for any $p, q \in Q_1$ and $a \in \Sigma$, as desired.

Let us now verify that $f(\tilde{M}_1)^\downarrow = f(\tilde{M}_1^\downarrow)$. By definition, $f(\tilde{M}_1) = (f(Q_1), \tilde{\Sigma}, \tilde{\delta}_2|_{f(Q_1) \times \tilde{\Sigma}}, q_{02}, F_2|_{f(Q_1)})$. We thus obtain that $f(\tilde{M}_1)^\downarrow = (f(Q_1), \Sigma, \delta_2^\downarrow|_{f(Q_1) \times \Sigma}, q_{02}, F_2|_{f(Q_1)})$ by a simple computation. On the other hand, since $\tilde{M}_1^\downarrow = (Q_1, \Sigma, \delta_1^\downarrow, q_{01}, F_1)$ and $f$ is a homomorphism from $\tilde{M}_1^\downarrow$ to $\tilde{M}_2^\downarrow$, we have that $f(\tilde{M}_1^\downarrow) = (f(Q_1), \Sigma, \delta_2^\downarrow|_{f(Q_1) \times \Sigma}, q_{02}, F_2|_{f(Q_1)})$, which is the same as $f(\tilde{M}_1)^\downarrow$. This completes the proof of 1). ∎

The equality $f(\tilde{M}_1)^\downarrow = f(\tilde{M}_1^\downarrow)$ (resp. $f(\tilde{M}_1)^\uparrow = f(\tilde{M}_1^\uparrow)$) in Proposition 5 means that the effect of acting $f$ on $\tilde{M}_1$ first and then retracting (resp. extending) is the same as that of first retracting $\tilde{M}_1$ (resp. extending) and then using $f$.

As an immediate consequence of Proposition 5, we have the following corollary.

*Corollary 1:* Keep the notations in Proposition 5. If there are homomorphisms $f : \tilde{M}_1 \longrightarrow \tilde{M}_2$ and $g : \tilde{M}_2 \longrightarrow \tilde{M}_1$, then $L(\tilde{M}_1^\downarrow) = L(\tilde{M}_2^\downarrow)$ and $L_w(\tilde{M}_1^\uparrow) = L_w(\tilde{M}_2^\uparrow)$.

*Proof:* It follows directly from Proposition 5 and Lemma 1. ∎

## VII. CONCLUSION

In this paper, we have introduced a direct, formal model of computing with words, where words are interpreted as fuzzy subsets of a finite set of symbols. We have related the new model to FACVs and FACAWs by establishing a retraction principle and a generalized extension principle, respectively. The retraction principle enables us to carry out computing with values via computing with words and the generalized extension principle enables us to carry out computing with all words via computing with some special words. Some relationships among retractions, extensions, and generalized extensions have been examined and their algebraic properties have also been investigated.

There are some problems which arise from the present formalization for computing with words and are worth further studying. A basic problem is how to choose words as the inputs of an FACW and how to rationally specify a fuzzy transition function such that the independence degree is as small as possible. In addition, it is worth noticing that much current interest in fuzzy control is devoted to model-based fuzzy control methods [22]. Recently, FACVs and fuzzy languages have been used to describe the so-called fuzzy discrete event systems, and the corresponding supervisory control theory has been developed [2], [3], [13], [20]. Some fuzzy automata with fuzzy inputs were applied to systems analysis in the

$$\begin{aligned}
\delta^\downarrow(p, wa_{n+1})(q) &= \delta^\downarrow(p, a_1 \cdots a_n a_{n+1})(q) \\
&= \cup_{q' \in Q}[\delta^\downarrow(p, a_1 \cdots a_n)(q') \cdot \delta^\downarrow(q', a_{n+1})(q)] \\
&= \vee_{q' \in Q}[\delta^\downarrow(p, a_1 \cdots a_n)(q') \wedge \delta^\downarrow(q', a_{n+1})(q)] \\
&= \vee_{q' \in Q}\Big\{\big[\vee_{A_1,\ldots,A_n \in \tilde{\Sigma}} [\tilde{\delta}(p, A_1 \cdots A_n)(q') \wedge \alpha]\big] \wedge \delta^\downarrow(q', a_{n+1})(q)\Big\} \\
&= \vee_{q' \in Q}\Big\{\vee_{A_1,\ldots,A_n \in \tilde{\Sigma}} \big[\tilde{\delta}(p, A_1 \cdots A_n)(q') \wedge \delta^\downarrow(q', a_{n+1})(q) \wedge \alpha\big]\Big\} \\
&= \vee_{q' \in Q}\Big\{\vee_{A_1,\ldots,A_n \in \tilde{\Sigma}} \big[\tilde{\delta}(p, A_1 \cdots A_n)(q') \wedge [\vee_{A_{n+1} \in \tilde{\Sigma}}[A_{n+1}(a_{n+1}) \wedge \tilde{\delta}(q', A_{n+1})(q)]] \wedge \alpha\big]\Big\} \\
&= \vee_{q' \in Q}\Big\{\vee_{A_1,\ldots,A_n \in \tilde{\Sigma}} \big[\vee_{A_{n+1} \in \tilde{\Sigma}} [\tilde{\delta}(p, A_1 \cdots A_n)(q') \wedge \tilde{\delta}(q', A_{n+1})(q) \wedge \alpha \wedge A_{n+1}(a_{n+1})]\big]\Big\} \\
&= \vee_{q' \in Q}\Big\{\vee_{A_1,\ldots,A_n,A_{n+1} \in \tilde{\Sigma}} \big[\tilde{\delta}(p, A_1 \cdots A_n)(q') \wedge \tilde{\delta}(q', A_{n+1})(q) \wedge \alpha \wedge A_{n+1}(a_{n+1})\big]\Big\} \\
&= \vee_{A_1,\ldots,A_n,A_{n+1} \in \tilde{\Sigma}}\Big\{\vee_{q' \in Q} \big[\tilde{\delta}(p, A_1 \cdots A_n)(q') \wedge \tilde{\delta}(q', A_{n+1})(q) \wedge \alpha \wedge A_{n+1}(a_{n+1})\big]\Big\} \\
&= \vee_{A_1,\ldots,A_n,A_{n+1} \in \tilde{\Sigma}}\Big\{\big[\vee_{q' \in Q} [\tilde{\delta}(p, A_1 \cdots A_n)(q') \wedge \tilde{\delta}(q', A_{n+1})(q)]\big] \wedge \alpha \wedge A_{n+1}(a_{n+1})\Big\} \\
&= \vee_{A_1,\ldots,A_n,A_{n+1} \in \tilde{\Sigma}}\big[\tilde{\delta}(p, A_1 \cdots A_n A_{n+1})(q) \wedge A_1(a_1) \wedge \cdots \wedge A_n(a_n) \wedge A_{n+1}(a_{n+1})\big].
\end{aligned}$$

---

$$\begin{aligned}
L(\tilde{M}^\downarrow)(w) &= \text{height}(\delta^\downarrow(q_0, a_1 \cdots a_n) \cap F) \\
&= \vee_{q \in Q}\{\delta^\downarrow(q_0, a_1 \cdots a_n)(q) \wedge F(q)\} \\
&= \vee_{q \in Q}\Big\{\big[\vee_{A_1,\ldots,A_n \in \tilde{\Sigma}} [\tilde{\delta}(q_0, A_1 \cdots A_n)(q) \wedge A_1(a_1) \wedge \cdots \wedge A_n(a_n)]\big] \wedge F(q)\Big\} \\
&= \vee_{q \in Q}\big[\vee_{A_1,\ldots,A_n \in \tilde{\Sigma}} [\tilde{\delta}(q_0, A_1 \cdots A_n)(q) \wedge A_1(a_1) \wedge \cdots \wedge A_n(a_n) \wedge F(q)]\big] \\
&= \vee_{A_1,\ldots,A_n \in \tilde{\Sigma}}\big[\vee_{q \in Q} [\tilde{\delta}(q_0, A_1 \cdots A_n)(q) \wedge A_1(a_1) \wedge \cdots \wedge A_n(a_n) \wedge F(q)]\big] \\
&= \vee_{A_1,\ldots,A_n \in \tilde{\Sigma}}\big[\vee_{q \in Q} [\tilde{\delta}(q_0, A_1 \cdots A_n)(q) \wedge F(q)] \wedge A_1(a_1) \wedge \cdots \wedge A_n(a_n)\big] \\
&= \vee_{A_1,\ldots,A_n \in \tilde{\Sigma}}\big[\text{height}(\tilde{\delta}(q_0, A_1 \cdots A_n) \cap F) \wedge A_1(a_1) \wedge \cdots \wedge A_n(a_n)\big] \\
&= \vee_{A_1,\ldots,A_n \in \tilde{\Sigma}}\big[L_w(\tilde{M})(A_1 \cdots A_n) \wedge A_1(a_1) \wedge \cdots \wedge A_n(a_n)\big].
\end{aligned}$$

---

early 1970s [16], [32]; a few attempts [4] and [8] have already been made for applying this kind of fuzzy automata to control. It might be highly interesting to bridge the gap among the supervisory control theory of FACVs, control of fuzzy automata with fuzzy inputs, and fuzzy control, by using our retraction principle and generalized extension principle.

ACKNOWLEDGMENT

The authors would like to thank the anonymous referees for some helpful suggestions.

APPENDIX I

To prove Theorem 1, it is convenient to have the following lemma.

*Lemma 2:* Let $\tilde{M} = (Q, \tilde{\Sigma}, \tilde{\delta}, q_0, F)$ be an FACW and $\tilde{M}^\downarrow = (Q, \Sigma, \delta^\downarrow, q_0, F)$ be the retraction of $\tilde{M}$. Then for any $p, q \in Q$ and $w = a_1 \cdots a_n \in \Sigma^*$, we have that

$$\delta^\downarrow(p, w)(q) = \vee_{A_1,\ldots,A_n \in \tilde{\Sigma}}\big[\tilde{\delta}(p, A_1 \cdots A_n)(q) \wedge A_1(a_1) \wedge \cdots \wedge A_n(a_n)\big].$$

*Proof:* We prove it by induction on $n$.

1) For the basis step, namely, $n = 0$, it is trivial.
2) The induction hypothesis is that the above equality holds for $w = a_1 \cdots a_n$. We now prove the same for $wa_{n+1}$, i.e., $a_1 \cdots a_n a_{n+1}$. Using the definition of $\delta^\downarrow$ and the induction hypothesis, we have the first equation shown at the top of the page, in which $A_1(a_1) \wedge \cdots \wedge A_n(a_n)$ is abbreviated to $\alpha$ for simplicity. This proves the lemma. ∎

*Proof of Theorem 1:* By the definition of $L_w(\tilde{M})$ and Lemma 2, we have the second equation shown at the top of the page, which completes the proof of the theorem. ∎

The idea of proving Theorem 2 is analogous to that of Theorem 1; let us first establish the following lemma.

*Lemma 3:* Let $\tilde{M} = (Q, \tilde{\Sigma}, \tilde{\delta}, q_0, F)$ be an FACW and $\tilde{M}^\uparrow = (Q, \mathcal{F}(\Sigma), \delta^\uparrow, q_0, F)$ be the generalized extension of $\tilde{M}$. Then for any $p, q \in Q$ and $W = A'_1 \cdots A'_n \in \mathcal{F}(\Sigma)^*$, we have that

$$\begin{aligned}
\delta^\uparrow(p, W)(q) = &\vee_{A_1,\ldots,A_n \in \tilde{\Sigma}} \vee_{a_1,\ldots,a_n \in \Sigma} \big[\tilde{\delta}(p, A_1 \cdots A_n)(q) \\
&\wedge A_1(a_1) \wedge \cdots \wedge A_n(a_n) \wedge A'_1(a_1) \wedge \cdots \wedge A'_n(a_n)\big].
\end{aligned}$$

*Proof:* We prove it by induction on $n$.

1) For the basis step, namely, $n = 0$, it is trivial.

$$\begin{aligned}
\delta^\uparrow(p, WA'_{n+1})(q) &= \delta^\uparrow(p, A'_1 \cdots A'_n A'_{n+1})(q) \\
&= \cup_{q' \in Q}[\delta^\uparrow(p, A'_1 \cdots A'_n)(q') \cdot \delta^\uparrow(q', A'_{n+1})](q) \\
&= \vee_{q' \in Q}[\delta^\uparrow(p, A'_1 \cdots A'_n)(q') \wedge \delta^\uparrow(q', A'_{n+1})(q)] \\
&= \vee_{q' \in Q}\left\{\left[\vee_{A_1,\ldots,A_n \in \tilde{\Sigma}} \vee_{a_1,\ldots,a_n \in \Sigma}[\tilde{\delta}(p, A_1 \cdots A_n)(q') \wedge \alpha \wedge \alpha']\right] \wedge \delta^\uparrow(q', A'_{n+1})(q)\right\} \\
&= \vee_{q' \in Q}\left\{\vee_{A_1,\ldots,A_n \in \tilde{\Sigma}} \vee_{a_1,\ldots,a_n \in \Sigma}[\tilde{\delta}(p, A_1 \cdots A_n)(q') \wedge \delta^\uparrow(q', A'_{n+1})(q) \wedge \alpha \wedge \alpha']\right\} \\
&= \vee_{q' \in Q}\left\{\vee_{A_1,\ldots,A_n \in \tilde{\Sigma}} \vee_{a_1,\ldots,a_n \in \Sigma}\left[\tilde{\delta}(p, A_1 \cdots A_n)(q') \wedge \left[\vee_{A_{n+1} \in \tilde{\Sigma}} \vee_{a_{n+1} \in \Sigma}[A_{n+1}(a_{n+1}) \right.\right.\right. \\
&\qquad\qquad\qquad\qquad\qquad\qquad \left.\left.\left. \wedge A'_{n+1}(a_{n+1}) \wedge \tilde{\delta}(q', A_{n+1})(q)]\right] \wedge \alpha \wedge \alpha'\right]\right\} \\
&= \vee_{q' \in Q}\left\{\vee_{A_1,\ldots,A_n \in \tilde{\Sigma}} \vee_{a_1,\ldots,a_n \in \Sigma}\left[\vee_{A_{n+1} \in \tilde{\Sigma}} \vee_{a_{n+1} \in \Sigma}[\tilde{\delta}(p, A_1 \cdots A_n)(q') \wedge \tilde{\delta}(q', A_{n+1})(q) \right.\right. \\
&\qquad\qquad\qquad\qquad\qquad\qquad \left.\left. \wedge \alpha \wedge A_{n+1}(a_{n+1}) \wedge \alpha' \wedge A'_{n+1}(a_{n+1})]\right]\right\} \\
&= \vee_{q' \in Q}\left\{\vee_{A_1,\ldots,A_{n+1} \in \tilde{\Sigma}} \vee_{a_1,\ldots,a_{n+1} \in \Sigma}\left[\tilde{\delta}(p, A_1 \cdots A_n)(q') \wedge \tilde{\delta}(q', A_{n+1})(q) \right.\right. \\
&\qquad\qquad\qquad\qquad\qquad\qquad \left.\left. \wedge \alpha \wedge A_{n+1}(a_{n+1}) \wedge \alpha' \wedge A'_{n+1}(a_{n+1})]\right\} \\
&= \vee_{A_1,\ldots,A_{n+1} \in \tilde{\Sigma}} \vee_{a_1,\ldots,a_{n+1} \in \Sigma}\left\{\vee_{q' \in Q}\left[\tilde{\delta}(p, A_1 \cdots A_n)(q') \wedge \tilde{\delta}(q', A_{n+1})(q) \right.\right. \\
&\qquad\qquad\qquad\qquad\qquad\qquad \left.\left. \wedge \alpha \wedge A_{n+1}(a_{n+1}) \wedge \alpha' \wedge A'_{n+1}(a_{n+1})]\right\} \\
&= \vee_{A_1,\ldots,A_{n+1} \in \tilde{\Sigma}} \vee_{a_1,\ldots,a_{n+1} \in \Sigma}\left\{\left[\vee_{q' \in Q}[\tilde{\delta}(p, A_1 \cdots A_n)(q') \wedge \tilde{\delta}(q', A_{n+1})(q)]\right] \right. \\
&\qquad\qquad\qquad\qquad\qquad\qquad \left. \wedge \alpha \wedge A_{n+1}(a_{n+1}) \wedge \alpha' \wedge A'_{n+1}(a_{n+1})\right\} \\
&= \vee_{A_1,\ldots,A_{n+1} \in \tilde{\Sigma}} \vee_{a_1,\ldots,a_{n+1} \in \Sigma}\left[\tilde{\delta}(p, A_1 \cdots A_{n+1})(q) \wedge \alpha \wedge A_{n+1}(a_{n+1}) \wedge \alpha' \wedge A'_{n+1}(a_{n+1})\right].
\end{aligned}$$

$$\begin{aligned}
L_w(\tilde{M}^\uparrow)(W) &= \text{height}(\delta^\uparrow(q_0, A'_1 \cdots A'_n) \cap F) \\
&= \vee_{q \in Q}\{\delta^\uparrow(q_0, A'_1 \cdots A'_n)(q) \wedge F(q)\} \\
&= \vee_{q \in Q}\left\{\left[\vee_{A_1,\ldots,A_n \in \tilde{\Sigma}} \vee_{a_1,\ldots,a_n \in \Sigma}[\tilde{\delta}(q_0, A_1 \cdots A_n)(q) \wedge \alpha \wedge \alpha']\right] \wedge F(q)\right\} \\
&= \vee_{q \in Q}\left[\vee_{A_1,\ldots,A_n \in \tilde{\Sigma}} \vee_{a_1,\ldots,a_n \in \Sigma}[\tilde{\delta}(q_0, A_1 \cdots A_n)(q) \wedge \alpha \wedge \alpha' \wedge F(q)]\right] \\
&= \vee_{A_1,\ldots,A_n \in \tilde{\Sigma}} \vee_{a_1,\ldots,a_n \in \Sigma}\left[\vee_{q \in Q}[\tilde{\delta}(q_0, A_1 \cdots A_n)(q) \wedge \alpha \wedge \alpha' \wedge F(q)]\right] \\
&= \vee_{A_1,\ldots,A_n \in \tilde{\Sigma}} \vee_{a_1,\ldots,a_n \in \Sigma}\left[\vee_{q \in Q}[\tilde{\delta}(q_0, A_1 \cdots A_n)(q) \wedge F(q)] \wedge \alpha \wedge \alpha'\right] \\
&= \vee_{A_1,\ldots,A_n \in \tilde{\Sigma}} \vee_{a_1,\ldots,a_n \in \Sigma}\left[\text{height}(\tilde{\delta}(q_0, A_1 \cdots A_n) \cap F) \wedge \alpha \wedge \alpha'\right] \\
&= \vee_{A_1,\ldots,A_n \in \tilde{\Sigma}} \vee_{a_1,\ldots,a_n \in \Sigma}\left[L_w(\tilde{M})(A_1 \cdots A_n) \wedge \alpha \wedge \alpha'\right].
\end{aligned}$$

2) The induction hypothesis is that the above equality holds for $W = A'_1 \cdots A'_n \in \mathcal{F}(\Sigma)^*$. We now prove the same for $WA'_{n+1}$, i.e., $A'_1 \cdots A'_n A'_{n+1}$. Using the definition of $\delta^\uparrow$ and the induction hypothesis, we have the first equation shown at the top of the page, where for convenience, we write $\alpha$ for $A_1(a_1) \wedge \cdots \wedge A_n(a_n)$ and $\alpha'$ for $A'_1(a_1) \wedge \cdots \wedge A'_n(a_n)$, respectively. This finishes the proof of the lemma. ■

*Proof of Theorem 2:* By the definition of word languages and Lemma 3, we have the second equation shown at the top of the page. Again, we write $\alpha$ for $A_1(a_1) \wedge \cdots \wedge A_n(a_n)$ and $\alpha'$ for $A'_1(a_1) \wedge \cdots \wedge A'_n(a_n)$, respectively. This proves the theorem. ■